\documentclass[letterpaper, 10 pt, conference]{ieeeconf}  

\IEEEoverridecommandlockouts                              

\usepackage{fancyhdr,graphicx,amsmath,amssymb}
\usepackage[ruled,vlined]{algorithm2e}

\usepackage{flushend}
\usepackage{verbatim}
\usepackage{booktabs}

\usepackage{dblfloatfix}

\usepackage{balance}
\usepackage{svg}
\usepackage{cite}

\title{\LARGE \bf
Task Allocation for Autonomous Machines using Computational Intelligence and Deep Reinforcement Learning
}
\author{Thanh Thi Nguyen$^{1}$, Quoc Viet Hung Nguyen$^{2}$, Jonathan Kua$^{3}$,\\ Imran Razzak$^{4}$, Dung Nguyen$^{5}$ and Saeid Nahavandi$^{6}$ 
\thanks{$^{1}$Faculty of Information Technology, Monash University, Australia {\tt\small thanh.nguyen9@monash.edu}}%
\thanks{$^{2}$Griffith University, Queensland, Australia}%
\thanks{$^{3}$Deakin University, Victoria, Australia}%
\thanks{$^{4}$Mohamed bin Zayed University of Artificial Intelligence, UAE}%
\thanks{$^{5}$The University of Queensland, Australia}%
\thanks{$^{6}$Swinburne University of Technology, Melbourne, Victoria, Australia {\tt\small snahavandi@swin.edu.au}}%
}

\begin{document}

\maketitle
\thispagestyle{empty}
\pagestyle{empty}

\begin{abstract}
Enabling multiple autonomous machines to perform reliably requires the development of efficient cooperative control algorithms. This paper presents a survey of algorithms that have been developed for controlling and coordinating autonomous machines in complex environments. We especially focus on task allocation methods using computational intelligence (CI) and deep reinforcement learning (RL). The advantages and disadvantages of the surveyed methods are analysed thoroughly. We also propose and discuss in detail various future research directions that shed light on how to improve existing algorithms or create new methods to enhance the employability and performance of autonomous machines in real-world applications. The findings indicate that CI and deep RL methods provide viable approaches to addressing complex task allocation problems in dynamic and uncertain environments. The recent development of deep RL has greatly contributed to the literature on controlling and coordinating autonomous machines, and it has become a growing trend in this area. It is envisaged that this paper will provide researchers and engineers with a comprehensive overview of progress in machine learning research related to autonomous machines. It also highlights underexplored areas, identifies emerging methodologies, and suggests new avenues for exploration in future research within this domain.

\end{abstract}

\section{INTRODUCTION}

Autonomous machines are increasingly being used across a wide range of fields, including manufacturing, transportation, agriculture, environmental monitoring, healthcare, retail, mining, and military applications \cite{huang2023learning,crosato2024social,zhang2024multi}. A useful autonomous system must be able to operate in various complex scenarios and capable of performing the assigned missions effectively and competently. When the mission complexity increases, it is desirable to share the burden with multiple vehicles or systems. The cooperation of multiple machines can simultaneously increase performance, decrease mission time, and enhance the mission success rate \cite{nguyen2020deep,li2025real}.  

Several challenges arise from the operation of multiple machines \cite{nguyen2019multi1,miele2025distributed}. These challenges are associated directly with \textit{task allocation} (TA) and coordination of systems of unmanned vehicles, which is significantly complicated when the numbers of vehicles is large. The purpose of TA is to assign a series of important and necessary tasks, e.g., reconnaissance, search, attack, and verification, to intelligent agents so as to maximise the overall mission performance. An effective TA solution is critical for achieving a high degree of efficacy in mission planning. Generally, TA is a combinatorial optimization problem that finds the minimal cost solution between two separate sets: a set of agents $A=\{a_1,a_2,...,a_n\}$ and a set of tasks $T=\{t_1,t_2,...,t_m\}$. To complete each task, an agent requires a cost. Let us define $C_{ij}$ as the non-negative cost of allocating the $i$th agent for handling the $j$th task. The aim is to allocate each task to one agent so as to minimize the total cost of finishing all tasks. The total allocation cost is characterized by:
\begin{equation}
    \sum (X_{ij} C_{ij}) \;\; \textrm{for} \; i=1,2,...,n \; \textrm{and} \; j=1,2,...,m
\end{equation}
where $X_{ij}$ is a binary variable with a value of 1 if agent $a_i$ is assigned to task $t_j$ and a value of 0 otherwise. 
To achieve an efficient allocation, the total cost needs to be minimized, and therefore TA becomes an optimization problem, as follows: 
\begin{equation}
\begin{aligned}
& \textrm{find} \; X_{ij} \\
& \textrm{subject to} \; \min \sum(X_{ij} C_{ij}) \; \\
& \qquad \qquad \qquad \textrm{for} \; i=1,2,…,n \; \textrm{and} \; j=1,2,…,m
\end{aligned}
\end{equation}

\begin{figure}[bp]
\centering
\includegraphics[width=3.0in]{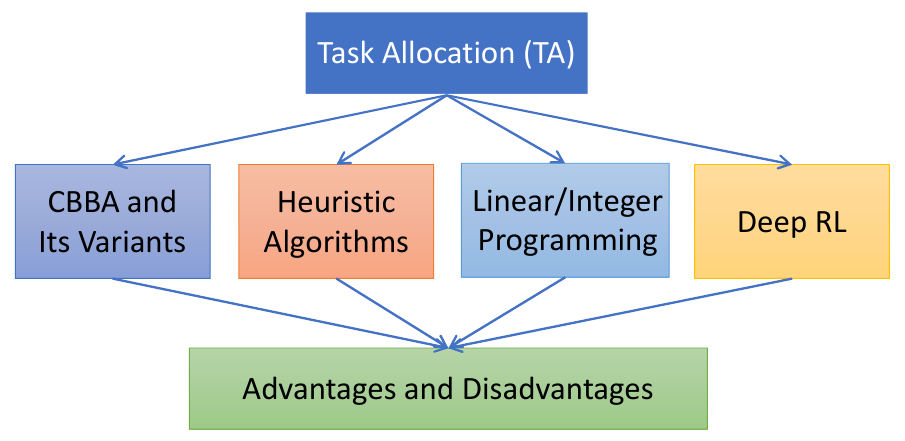}
\caption{A taxonomy of TA methods included in this review. CBBA is the abbreviation of consensus-based bundle algorithm while RL stands for reinforcement learning.}
\label{fig_taxonomy}
\end{figure}

The existing work in Seenu et al. \cite{seenu2020review} reviews TA methods for multiple-robot systems, but it does not cover the latest developments in this area, especially the emerging applications of deep reinforcement learning (RL) methods. In this paper, a comprehensive literature review covering computational intelligence and deep RL methods for TA is conducted. Fig. \ref{fig_taxonomy} presents the structure of the survey and taxonomy of reviewed techniques. The benefits and drawbacks of the surveyed methodologies are analysed and contrasted thoroughly.

\section{TASK ALLOCATION: A SURVEY}
The TA methodologies can be categorised into two groups, i.e., centralised and decentralised methods \cite{oh2015market}. 
In \textit{centralised} TA methods, a single control center obtains the entire information from all agents, and then processes and directs suitable commands to each agent. Linear/integer programming-based techniques are examples of centralized methods that attempt to acquire an optimal but computationally inexpensive solution \cite{radmanesh2016flight}. With regard to evolutionary approaches, genetic algorithm (GA), which can offer sub-optimal solutions with a reduced computational burden, is a popular method for centralised TA problems \cite{darrah2013using}. Similarly, particle swarm optimization (PSO) has also been an efficient method for centralized TA problems \cite{wang2013applied, li2016modified}.

In \textit{decentralized} TA methods, each agent acquires local information by itself, or through communications with neighboring agents. Therefore, it can cope with dynamic and unanticipated situations more reliably. In addition, decentralised TA methods can mitigate the computational burden by aggregating the computing power from each constituent agent. Nevertheless, the consensus on situation awareness poses a key challenge in decentralised TA methods. 
Market-based techniques have been proposed to address this challenge in a decentralized system. Each participant in the virtual market gives a decision based on its own profit and circumstances, therefore increasing the effectiveness of every participant. A market mechanism such as an auction is used to distribute resources to the participants in this virtual market \cite{oh2015market}. 

\begin{figure}[htp]
\centering
\includegraphics[width=3.4in]{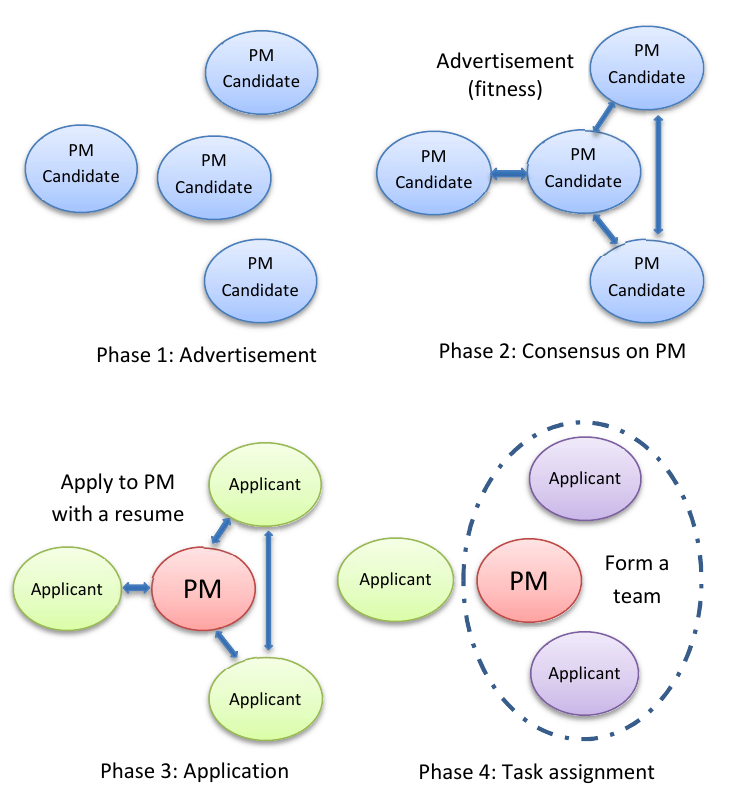}
\caption{Four phases of the TA process in a virtual market \cite{oh2015market}. The project manager (PM) is the candidate who has the highest fitness score.}
\label{fig_virtual_market}
\end{figure}

Fig. \ref{fig_virtual_market} illustrates the concept of the virtual market and its decentralised selection procedure. Agents disclose their information to others, in order to build a team and to acquire a task, which is known as a project. In the first phase, every agent can advertise regardless of the number of neighbor candidates (Phase 1). The suitability of a candidate to complete a specific task is measured by a fitness function. A candidate is selected as a project manager if it has the highest fitness score (Phase 2). As soon as the manager is determined, other agents send an application resume to the manager including estimated information on completing the applied task (Phase 3). A consensus on the application letters is made among agents by sharing their application letters with their neighborhoods. Using this method, distant agents do not need to use a network topology to reach the manager. Based on the resumes, the manager chooses a number of members among the applicants to form a team (Phase 4). This procedure is repeated to assign a specific task to the team members, and at the end, each agent acquires its own task. A polynomial-time algorithm combining both \textit{auction} and \textit{consensus} methods, known as the consensus-based bundle algorithm (CBBA), was introduced in \cite{choi2009consensus}, and detailed in the following subsection.

\subsection{CBBA and its variants}
Cooperation in an agent fleet is crucial for improving the overall performance in accomplishing a mission. Centralized methods communicate situation awareness to a centralized control server that produces a plan for the whole fleet. These kinds of systems have the advantage that they put much of the processing burden securely on the ground, allowing the agents (machines) to be smaller and more cost-effective to build. However, agents need to constantly communicate with the fixed server, therefore decreasing the mission handling power of the fleet, as well as generating a single point of control, which is vulnerable to failure in the system.

To increase the ability to handle a large mission range and also to eliminate the single point of failure, decentralized methods equip each agent with a centralized planner. In case of limited communication, which is not uncommon in realistic networks, inconsistencies in situation awareness might lead to conflicting TA because agents obtain optimal solutions based on different sets of information. As a result, \textit{consensus algorithms} are often used to formulate a consistent situation awareness before TA takes place. These consensus algorithms make sure that situation awareness is converged over different network topologies, which allows the system to perform TA in highly dynamic and uncertain situations. The consensus algorithms have disadvantages as they take a considerable amount of time to converge to a consistent situation awareness condition, and often require transmitting a large amount of data. This can lead to severe delays in low-bandwidth environments, and considerably increase the total TA time for the fleet.

Another class of methods for TA is \textit{auction algorithms}. Generally, agents place bids on tasks with values using their own situation awareness, and the greatest bid wins the allocation. Each task is to be allocated to a single agent as only one agent is chosen as the winner. Therefore, convergence to a conflict-free solution can be reached for most auction algorithms even agents may have inconsistent situation awareness. The shortcoming of these algorithms is that the bids from the agents must be transferred to the auctioneer. The types of network topologies are therefore limited because a connected network is needed among the agents to convey all the bid information. 

Consensus methods are generally robust to network topologies, whereas traditional auction methods are computationally effective and robust to inconsistencies in situation awareness. To take their advantages, a combination of these two methods, namely CBBA, was introduced in \cite{choi2009consensus}.

The consensus-based grouping algorithm (CBGA) proposed in \cite{hunt2012extension1,hunt2014consensus} extends CBBA by introducing a new data storage system and a multiagent consensus algorithm to handle tasks that need more than one agent to complete. CBGA can handle multiagent multi-TA with multiple agent requirements pertaining to UAVs. 
Conventional CBBA assumes homogeneity of agents, which is often violated in real practice. Heterogeneous robots consensus-based allocation (HRCA) suggested in \cite{di2011consensus} extends CBBA by handling constraint violations in the second stage. The agents (robots) use an iterative procedure to re-distribute the tasks that exceed their individual capability with a minimal loss in terms of the score function. Therefore, HRCA is able to handle randomly generated heterogeneous robotic networks \cite{di2011consensus}.

\subsection{Heuristic algorithms for TA problems}
Evolutionary algorithms are generic metaheuristic techniques that perform optimization based on a population of evolutionary individuals. They employ mechanisms inspired by biological evolutionary processes to discover optimal or near-optimal solutions for complex and sometimes NP-hard problems. These mechanisms include reproduction, mutation, recombination, and selection. One of the key evolutionary algorithms is GA, which was utilized in numerous applications \cite{khatami2019ga, tousi2024genetic, klampfl2024using}. Other heuristics that have been broadly employed for TA include swarm-based algorithms, such as PSO.

GA has a number of advantages for undertaking TA problems. It does not require explicit calculation of the gradient of the cost function and can handle nonlinear cost functions. GA is able to produce solutions that are significantly better than those from the random search method and it has a possibility of parallel implementation \cite{darrah2013using}.

On the one hand, PSO \cite{kang2011novel,yin2006hybrid,li2013attractor} runs faster than GA in TA. This is because PSO attempts to balance exploration (through neighboring experience) and exploitation (through self-experience) by combining global and local search methods, respectively. As such, it has a great ability to escape from local optima. On the other hand, a variant of PSO, i.e., discrete PSO \cite{kang2011novel}, performs better than PSO, GA, simulated annealing, and tabu search in terms of search quality to expedite convergence by embedding a variable neighborhood descent search method based on a transfer and swap neighborhood. Good performance of discrete PSO also results from a new migration mechanism to strike an equilibrium between extensive exploration and targeted exploitation \cite{kang2011novel}.

In case of large, self-operating, multi-robot systems, scalability and the capability of adapting to different environments are the utmost importance feature of TA algorithms. Based on the bee algorithms \cite{ozbakir2010bees,tapkan2013solving}, a distributed bee algorithm was proposed in \cite{jevtic2011distributed,jevtic2012distributed} for assigning robots to different positions in a 2-D arena. The qualities of targets, which are characterized by scalar values, are used to derive the expected distribution. The decision-making system is distributed, and the distances and qualities of targets are taken into account so that the robots can autonomously obtain their allocated tasks. The distributed bee algorithm provides a robot swarm with not only the flexibility where the total of robots and targets can be varied but also the ability to adapt to a non-uniform distribution of the attributes of targets \cite{jevtic2012distributed}.

Many studies have focused on the TA problem of homogeneous UAVs, but the task diversity requires a higher capability of a UAV system. A modified GA for TA was proposed in \cite{han2021modified} for a set of heterogeneous UAVs with limited resource constraints. They introduced the idea of a fuzzy elite degree to enhance gene retention and intensity in GA operations. For example, in the GA selection process, the fuzzy elite degree of the $i$th individual is defined by:
\begin{equation}
\begin{aligned}
    E_i = \alpha F_i + \beta \Delta_i
\end{aligned}
\end{equation}
where $\alpha$ and $\beta$ are coefficients with $\alpha+\beta=1$. These coefficients respectively control the influence of $F_i$ and $\Delta_i$ with their definitions as follows:
\begin{equation}
\begin{aligned}
    F_i = \frac{\log_2 \left(f_i/\max_i (f_i)+1 \right)}{\max_i \left( \log_2 \left( f_i/\max_i (f_i) + 1 \right) \right)}
\end{aligned}
\end{equation}
and
\begin{equation}
\begin{aligned}
    \Delta_i = \frac{\max_i \left( 1 + \exp(10*(A_i/\max_i(A_i)-l_s)) \right)}{1+\exp(10*(A_i/\max_i(A_i)-l_s))}
\end{aligned}
\end{equation}
where $f_i$ and $A_i$ are the fitness score and the number of times the $i$th individual appears in the current iteration, and $l_s$ is used to adjust the range of $A_i$. Accordingly, individuals having higher elite degrees will be chosen as elite ones. It means that individuals with high fitness value $f$ are selected with high probability, but individuals with few $A$ also have a chance to be chosen to maintain diversity within the population. Experimental results on a complex TA problem of a heterogeneous UAV system demonstrate an improved searchability and convergence performance of this approach compared with its competing methods.

\subsection{Integer/linear programming (LP) for TA}

While CBBA (and its variants) and heuristic approaches have shown remarkable performance in complex TA problems, the use of classical integer/LP methods is still growing. 
For example, a mission planning problem for military aircraft was considered in \cite{quttineh2013military} in which a given fleet of aircraft attacks several ground targets. The airspace is discretised such that the strike positions and lighting positions can be characterized by nodes. Aircraft movements are represented by arcs, which are defined by a priori knowledge. The goal is to optimize the outcome and mission timespan of the attacks. The problem is modeled as an expanded vehicle routing problem where timing and precedence constraints are taken into account. A mixed-integer linear programming is then used to solve the problem.

In numerous practical situations, robot fleets must address not only delivery requests but also additional factors like energy efficiency and the need to avoid human-centered workspaces (e.g., avoiding areas with high foot traffic). The study in \cite{wilde2024statistically} framed the multi-robot pickup and delivery problem as a multiobjective optimization challenge, aiming to identify policies that balance various tradeoffs among these objectives. A key aspect of that problem is accounting for the variability in objective values due to the unpredictable nature of task arrivals. That study introduces an adaptive sampling method designed to identify policies that are nearly optimal, approximate the full range of most effective solutions, and are statistically different from one another. The method demonstrates clear advantages over baselines and shows robustness through sensitivity analysis.

TA and safety are two critical challenges in collaborative robotics because they involve the interaction between two key resources, the operator and the collaborative robot (cobot), to achieve optimal performance while maintaining a safe workspace. However, most methods tend to prioritize one of these issues over the other. To address this gap, a model based on integer programming was proposed in \cite{faccio2024task} for collaborative assembly systems, incorporating safety constraints with the goal of minimizing the total task completion time. The innovation of this model lies in its consideration of the distance between resources and its ability to adjust the cobot's speed accordingly. By allowing the cobot to maintain higher speeds for extended periods, the model ensures that the safety constraint is met by maximizing the time resources remain at a safe distance from each other. This approach effectively balances performance and safety in collaborative settings. Although experimental results validate and underscore the necessity of the proposed method, one of its limitations is that it does not account for stochastic times or precedence constraints, which could restrict the range of possible solutions.

\subsection{Deep Reinforcement Learning (RL) for TA}

\subsubsection{Deep RL Background}
RL is a class of machine learning methods, which are the backbone of many sequential decision-making support systems. RL tasks are normally characterised by a Markov decision process (MDP), which establishes a mathematical framework for analyzing sequential decision-making problems \cite{goel2025unveiling}. An MDP is a tuple $\langle  \mathcal{S},\mathcal{A},\mathcal{P},\mathcal{R},\gamma \rangle$ where $\mathcal{S}$ is a collection of states, $\mathcal{A}$ is a collection of actions, $\mathcal{P}$ represents a matrix of state transition probabilities $\mathcal{P}_{ss'}=\mathbb{P}[S_{t+1}=s'|S_t=s,A_t=a]$, $\mathcal{R}$ signifies a reward function $\mathcal{R}_s=\mathbb{E}[R_{t+1}|S_t=s,A_t=a]$, and $\gamma$ is a discount factor $\gamma \in [0,1]$. The behaviors of an agent can be defined by a policy $\pi$, which is the likelihood of taking an action given a state:
\begin{equation}
    \pi(a|s)=\mathbb{P}[A_t=a|S_t=s]
\end{equation}

To find an optimal policy, we strive to optimize the expected return, $\mathbb{E}[G_t]$, at each step $t$, where $G_t$ is the total discounted reward:
\begin{equation}
\begin{split}
G_t & =R_{t+1}+\gamma R_{t+2}+\gamma^2 R_{t+3}+\dots \\ & =\sum_{k=0}^{\infty}\gamma^kR_{t+k+1}
\end{split}
\end{equation}
Optimal policies, denoted as $\pi_*$, are the ones that are superior to or at least equivalent to all the others. These optimal policies feature similar optimal state-value $v_*(s)=\max_{\pi} v_\pi(s)$ for all $s\in \mathcal{S}$ where $v_\pi(s)$ is the state-value function of an MDP, which is the estimated return by beginning with state $s$, and then pursuing policy $\pi$:
\begin{equation}
\begin{split}
    v_\pi(s)&=\mathbb{E}_\pi[G_t|S_t=s]\\&=\mathbb{E}_\pi[R_{t+1}+\gamma v_\pi(S_{t+1})|S_t=s]
\end{split}    
\end{equation}

Complex problems normally have a large number of states, which hinder algorithms such as dynamic programming to find direct solutions. Approximate solutions by RL methods are therefore advocated. For high-dimensional problems where state and action spaces are large, conventional RL methods may struggle to learn optimal policies effectively. Deep learning, known for its ability to solve high-dimensional problems, has been integrated into RL methods to form what is called deep RL. Deep RL therefore becomes an excellent tool for applications where decisions must be made in large-scale environments. Deep RL supports ongoing learning and can create agents that are more reliable and scalable than their human counterparts \cite{nguyen2019manipulating}.

\subsubsection{Applications of Deep RL for TA}

In a multiagent system, an agent is normally unable to obtain information or observe the behaviors of other agents. To perform complex tasks cooperatively, agents can learn to communicate via a communication channel before taking actions. A TA method based on deep RL was introduced in \cite{noureddine2017multi}, namely cooperative deep Q-learning, to improve the communication and social cooperation between agents based on the communication neural net \cite{sukhbaatar2016learning}. Three types of agents are defined, including manager, participant, and mediator. The manager is an agent that can request help from other agents to perform its task while the participant is an agent that accepts and carries out the declared task. The mediator is an agent that commits to find participants for another agent. An agent can be in one of the three states, i.e., busy, committed, or idle, at any time step. The manager or participant agent will have the busy state while the mediator agent will have the committed state. An agent that is free and not allocated or committed to any task will have the idle state and only this kind of agent can be allocated a new task. 

Once the complex problem is broken down into several tasks, the manager agent will send messages to all of its neighbors and idle agents will submit bids to perform the tasks. Depending on the resources of these neighbors, the manager will form a full group or a partial group for the announced task. In case of a partial group, the manager can commit an unfulfilled task to a mediator agent, which is one of the neighboring agents. An agent is more likely to be selected as a mediator if it has more neighbors. This is a repeated process until the resources required for the announced task are met. Experimental results by using different types of social networks as in \cite{de2007distributed} demonstrate a great performance of the proposed cooperative deep RL approach. The agents are able to learn not only to allocate tasks efficiently but also to prepare properly for new tasks. 

The development of wireless technology, e.g., the 5G network, has enabled the continuing enhancement of the Internet of Vehicles, and that has attracted more and more users. Mobile edge computing (MEC) has provided the edge server infrastructure to facilitate the data processing and transmission tasks to improve the quality of experience of vehicle users. Given the increasing demand for multimedia services from vehicle users and the fast-moving speed of vehicles, it is still challenging to ensure efficient and stable data transmission to users. A pre-caching and TA approach based on deep RL was proposed in \cite{ma2020deep} to address this problem in a simulation environment consisting of multiple roadside units and vehicles. 
The RL algorithm is used to learn an intelligent policy to cache data and allocate the data transmission tasks collaboratively between the vehicle to roadside unit and vehicle to vehicle communication modes. Simulation results exhibit the efficiency of the RL method and its superiority against competing policies in terms of maximizing the data receiving rate of fast-moving vehicles.

With the low-cost and strong mobility capabilities, UAVs have been used to improve the connectivity in the MEC technology. A traditional MEC normally has edge servers in fixed cellular base stations, which can negatively affect the service quality provided to mobile users. Multiple UAVs can be used as mobile servers in the MEC to enhance the computational offloading services to mobile devices. However, it is challenging to solve the TA problem for UAVs in this kind of UAV-assisted MEC system because of its non-convex characteristics. Several studies have attempted to solve this problem using greedy algorithm \cite{wang2019joint}, Lyapunov optimization method \cite{wang2020task}, Q-learning \cite{elgendy2021joint}, and iterative algorithm \cite{he2021joint}, but they have limitations in handling action spaces that are high-dimensional and continuous. On the other hand, this problem was formulated in \cite{yu2021deep} as an MDP and used the twin delayed deep deterministic policy gradient proposed in \cite{fujimoto2018addressing} to find optimal solutions in a continuous action space. Each UAV is characterized by a deep RL agent implemented by the twin delayed deep deterministic RL algorithm \cite{lillicrap2016continuous}. 

On the other hand, a multiagent RL method was designed in \cite{zhou2024multirobot} to dynamically schedule tasks in cooperative multi-robot settings, considering the service status of industrial robots. The approach establishes constraints that define the interactions between tasks and robots and develops serviceability evaluation and error compensation agents within the deep deterministic policy gradient RL framework to evaluate robot serviceability and correct positioning errors according to their operational status. Furthermore, a heuristic graph convolution scheduling agent is created by integrating heuristic rules with graph CNNs to assign tasks according to the characteristics of both tasks and robots and their interdependencies. The approach facilitates both static TA and adaptive scheduling in multi-robot collaborations, effectively addressing the single-task, multi-robot, and time-extended assignment problem in industrial settings.

Design and planning for a multi-robot system often involve two basic steps, i.e., TA and task execution. Most approaches are for either TA or path planning. An end-to-end deep RL approach to both multi-robot TA and path planning was proposed in \cite{elfakharany2021end} based on the deep RL proximal policy optimization method \cite{schulman2017proximal}. The experiments are performed on a 2D plane environment having both fixed and moving obstacles. The robots are homogeneous and the number of robots is equal to the number of goal locations, i.e., targets. Each robot is encouraged to select and approach a specific target and aim to minimize its moving time to the target. The robot is penalized if it collides with obstacles or other robots or reaches the same target as other robots. The {TurtleBot~3 Waffle Pi} robots \cite{ROBOTIS} are used for experiments with the training and testing based on the Gazebo simulator. The results show better performance regarding the success rate, moving time, and distance of the proposed method compared with methods focused solely on navigation.

\section{PROS AND CONS OF TA METHODS}
The advantages of consensus-based methods, i.e., CBBA and its variants, stem from the integration of market-based auction strategy and consensus method \cite{brunet2008consensus,argyle2011multi,mu2013value,oh2014market}. The auction approach is used as a framework for decentralized TA while the consensus method based on local communication is applied to resolve conflicts and to reach an agreement on the winning bid. Therefore, CBBA and its variants have better performance and faster convergence speed as compared with the established auction-based TA methods~\cite{choi2009consensus}.

CBBA and its variants however have a number of drawbacks as pointed out in the literature when applied to complex TA problems \cite{choi2009consensus,hunt2012extension1,hunt2014consensus,bertuccelli2009real,mercker2010extension}. 
CBBA does not facilitate the assignment of tasks with precedence requirements \cite{mercker2010extension}. CBBA requires a multiagent system to know the global presence of a task. When the agents search their areas of responsibility and find new targets, the presence of these new targets or tasks must be shared with each agent, and the procedure needs to be re-run with the newly acquired information \cite{mercker2010extension}.

More importantly, CBBA can only give conflict-free solutions with a guaranteed 50\% optimality when solving multiagent multi-TA problems \cite{choi2009consensus,hunt2012extension1,hunt2014consensus}. It was highlighted in \cite{bertuccelli2009real} that CBBA does not explicitly possess the ability to address obstacle avoidance and its allocation results may be highly susceptible to input noise and can produce churning behaviors. Alternatively, CBBA can only handle homogeneous robotic systems in which robots are assumed to have similar characteristics and capabilities as pointed out in \cite{di2011consensus}.

The extensions of CBBA, e.g., extended CBBA \cite{mercker2010extension} or CBGA \cite{hunt2012extension1,hunt2014consensus} have demonstrated that they are able to overcome the shortcomings of CBBA, and can be applied to complex scenarios. HRCA can be applied for TA of UAVs where the networks of UAVs are heterogeneous, i.e., UAVs are not identical owing to their sensing or actuation hardware and features \cite{di2011consensus}. 

With regard to the evolutionary algorithms, they have been found to be increasingly used for TA. GA is one of several traditional methods in this category. Although GA has been widely used for addressing TA problems \cite{darrah2013using}, it has a significant disadvantage related to its computational expense \cite{yin2006hybrid}. GA also generates solutions with higher costs as compared to PSO \cite{yin2006hybrid}. Extensions of PSO, e.g., discrete PSO \cite{kang2011novel}, hybrid PSO \cite{yin2006hybrid}, and reverse predictor PSO \cite{li2016modified} show efficiency in terms of convergence speed and average costs. Other evolutionary algorithms developed for TA of robot swarms, e.g., distributed bee algorithm \cite{jevtic2012distributed} or processors in distributed systems, e.g., novel global harmony search algorithm \cite{zou2010novel}, have a good potential for applications to complex TA problems.

On the other hand, deep RL methods have attracted huge attention recently and their applications for TA have emerged. They are an end-to-end approach as they can take data directly from the sensors and learn a policy that is able to generate optimal sequential solutions for autonomous machines. Deep RL methods have a disadvantage that their training process may require significant computational resources because building a deep RL agent is only effective when it has a large amount of environment interactions~\cite{nguyen2023solving}. Furthermore, the input sensory data for deep RL algorithms are normally obtained in terms of high-dimensional images/signals of the environment. The training time of deep RL methods therefore may be greater than heuristics and evolutionary computation algorithms because of the time needed to process high-dimensional data and allow the RL agent to sufficiently engage with the environment. Taking the input environment via its sensory data representation however may be an advantage of deep RL methods because this kind of modeling allows more realistic features to be taken into account \cite{nguyen2019new}. With the power of representation learning, deep neural networks can learn and obtain a compact feature set from raw high-dimensional images or sensory data of the environment, which enables the network to approximate complex functions describing the relationship between inputs and action outputs. Deep RL methods therefore have made significant contributions in control and coordination of autonomous machines as reviewed in this study. 

\section{OPEN ISSUES AND FUTURE DIRECTIONS}

\subsection{Benchmarking TA methods}
There have been numerous TA methods reviewed in this study, but they are mostly evaluated using different environments and experiment settings. In each category, there are traditional methods with their improved versions still being developed over time, but there are also new approaches that benefit from recent machine learning advances such as deep RL. Each method has its own merits and demerits depending on its nature and the environment in which it is used. For example, a linear programming method can be very efficient in a simple TA problem, but its performance may not be comparable with that of an evolutionary computation method in a more sophisticated problem. In the TA literature, there is currently no comprehensive work on comparing variants of CBBA with linear/integer programming, GA, PSO, bat algorithm, bee algorithm, harmony search, and the emerging deep RL methods using a benchmark environment setting. Benchmarking different TA methods in an empirical study is therefore highly recommended as it would provide a comparative evaluation of these methods and give insights into the employability of each method in specific environments. This is a research direction that would promote broader applications of the existing TA methods and also facilitate the development of new and effective methods in this area.

\subsection{Enabling heterogeneous machines}
Most approaches proposed for TA have been dealing with machines that have the same or similar features and functions, i.e., homogeneous agents. Real-world problems normally have to cope with complex tasks requiring agents to be heterogeneous machines with different features and functions \cite{li2024efficient}. For example, UAVs operate in the air while robots perform their tasks on the ground. The tasks assigned for UAVs and robots therefore are dissimilar, and thus the existing methods may struggle to handle them. Recent studies have attempted to address this research topic, but they are still not to the full extent. For example, a multi-robot task and motion planning in a manufacturing environment was introduced in \cite{zheng2021parallel}, but their experiments are limited to two types of robots that have similar characteristics. Further studies in this direction are important to accommodate more variety of machines and thus maximize the applicability of autonomous machines to real-world problems.

\subsection{The IoT to facilitate interactions between machines}
In an environment with multiple autonomous machines, the interactions among the machines are important for efficient route planning and TA. The IoT technologies can be utilized to facilitate these interactions as IoT devices can be mounted on the machines and various static and dynamic objects. The IoT can enable autonomous machines to interact with their environment, recognize obstacles, and move through them. This would create a well-connected, dynamic, and ubiquitous environment with rich data sources including the position of objects, the existence of obstacles, and the changes occurring in the environment \cite{tashtoush2021enhancing}. While TA in such environments is more challenging, it also presents significant opportunities to enhance the performance of these algorithms. A future research in this direction is to incorporate the full extent of IoT into TA. IoT-enabled machines will be an emerging research trend in the coming years.

\begin{table}[!t]
\centering
\begin{footnotesize}
\caption{A summary of reviewed methods for TA}
\label{table1}
\begin{tabular}{l l l}
\toprule
\textbf{Algorithms} & \textbf{Surveyed Methods} \\
\midrule
Linear/integer programming
& \cite{quttineh2013military, wilde2024statistically, faccio2024task, karasakal2011branch, solovey2021fast, latini2024mixed} \\
\midrule
CBBA and its variants
& \cite{oh2015market, choi2009consensus, hunt2012extension1, hunt2014consensus, di2011consensus, brunet2008consensus, argyle2011multi, mu2013value, oh2014market, bertuccelli2009real, mercker2010extension} \\
\midrule
Genetic algorithm
& \cite{darrah2013using, han2021modified, martin2021multi, liau2022genetic,xidias2024balanced} \\
\midrule
PSO, ant colony, swarm intelligence
& \cite{wang2013applied, li2016modified, kang2011novel, li2013attractor, wang2011discrete, wang2023collaborative, amorim2020assessing} \\
\midrule
Bees/bat algorithm, harmony search
& \cite{ozbakir2010bees, tapkan2013solving, jevtic2011distributed, jevtic2012distributed, zou2010novel, del2012centralized, umlauft2023bees} \\
\midrule
Deep reinforcement learning
& \cite{ma2020deep, yu2021deep, zhou2024multirobot, elfakharany2021end, lin2020blockchain, zhao2019fast, chang2020coactive} \\
\bottomrule
\end{tabular}
\end{footnotesize}
\end{table}

\section{CONCLUSIONS}
It is difficult for humans to perform mission planning with large systems of machines. Artificial intelligence and machine learning have been applied to automate this process. Collaboration and sharing of information among vehicles are required to maximise the effectiveness when performing missions involving multiple vehicles. In this paper, we have presented a literature review of computational intelligence and deep RL methodologies for TA in changing environments. A summary of the reviewed methods is presented in Table~\ref{table1}. Each class of methods however has its respective advantages and disadvantages, as discussed in this paper. While computational intelligence and deep RL methods constitute a great direction to tackle TA problems for autonomous machines, further in-depth research is required to properly engage the related methodologies. We have suggested several interesting research directions, which are worth investigating as they will certainly contribute to improving the efficiency and effectiveness of machine learning and artificial intelligence methods for controlling and coordinating autonomous machines in complex and dynamic real-world problems.


\bibliographystyle{IEEEtran}
\bibliography{references}

\end{document}